\documentclass{article}
\usepackage{spconf,amsmath,graphicx,hyperref}
\usepackage{booktabs} 
\usepackage{amssymb}
\usepackage{xcolor}
\usepackage{etoolbox}
\apptocmd{\thebibliography}{\setlength{\itemsep}{1.41pt}}{}{}
\usepackage[table]{xcolor}
\newcommand{\blank}{} 
\definecolor{oursblue}{RGB}{230,240,255}
\definecolor{lightgray}{gray}{0.95}

\title{MOTION-GUIDED SEMANTIC ALIGNMENT WITH NEGATIVE PROMPTS FOR ZERO-SHOT VIDEO ACTION RECOGNITION}
\name{Yiming Wang, Frederick W. B. Li, Jingyun Wang\thanks{This work has used Durham University’s NCC cluster. NCC has been purchased through Durham University’s strategic investment funds, and is installed and maintained by the Department of Computer Science.}}
\address{Department of Computer Science, Durham University}
\begin{document}
\ninept
\maketitle
\begin{abstract}
Zero-shot action recognition is challenging due to the semantic gap between seen and unseen classes. We present a novel framework that enhances CLIP with disentangled embeddings and semantic-guided interaction. A Motion Separation Module (MSM) separates motion-sensitive and global-static features, while a Motion Aggregation Block (MAB) employs gated cross-attention to refine motion representation without re-coupling redundant information. To facilitate generalization to unseen categories, we enforce semantic alignment between video features and textual representations by aligning projected embeddings with positive textual prompts, while leveraging negative prompts to explicitly model “non-class” semantics. Experiments on standard benchmarks demonstrate that our method consistently outperforms prior CLIP-based approaches, achieving robust zero-shot action recognition across both coarse and fine-grained datasets.
\end{abstract}
\vspace{-5pt}
\section{Introduction}
\label{sec:intro}
\vspace{-5pt}
In recent years, large-scale vision–language pre-trained models such as CLIP~\cite{radford2021learning} have shown remarkable success in cross-modal learning, driving significant advances in zero-shot learning (ZSL)~\cite{palatucci2009zero}. Extending this paradigm to the video domain, zero-shot action recognition (ZSAR)~\cite{chen2021elaborative} seeks to classify unseen actions by transferring knowledge from seen categories through a shared visual–textual space. Unlike static image classification, video understanding requires modeling long-range temporal dynamics and fine-grained motion cues while mitigating background interference, making generalization to novel classes considerably more challenging.

Early efforts such as JigSawNet~\cite{le2019jigsawnet} explored connections between visual and textual spaces through various mechanisms, laying the groundwork for subsequent vision–language approaches. Building on the success of CLIP, IV-L~\cite{ju2022prompting} first adapted it to videos via text prompts, which improved generalization but only marginally. ActionCLIP~\cite{wang2023actionclip} advanced this direction with a temporal aggregation module, yet its performance remained sensitive to background bias and limited by the lack of fine-grained temporal modeling. 
In recent years, research on adapting CLIP for video zero-shot action recognition has evolved along three main directions. Prompt-based methods~\cite{wasim2023vita,ahmad2023ez,gowda2025temporal,jingyi2025kronecker,liang2025zar} introduce learnable or dynamic prompts to better align CLIP with video inputs, ranging from temporal visual prompts and semantic action decomposition to test-time adaptive tuning. Textual enrichment approaches~\cite{lin2023match,gowda2024telling,yu2025building} enhance semantic diversity through large language models, story-based embeddings, or knowledge-graph-guided augmentation, but often risk introducing background noise or irrelevant information. Representation alignment strategies~\cite{huang2024froster,gowda2024continual} aim to maintain CLIP's semantic space via techniques such as residual distillation and continual learning, but they do not explicitly capture temporal dynamics.

From prior work, we observe that visual prompt tuning often involves larger parameter overhead and more complex designs, while textual prompts are mostly restricted to positive descriptions, which can lead to ambiguity in fine-grained action categories where multiple expressions may exist. Moreover, although these methods have advanced ZSAR, they remain constrained by insufficient temporal modeling and semantic noise in cross-modal alignment. This motivates the need for a framework that explicitly disentangles motion from static context while ensuring reliable semantic alignment.

In view of the above discussion, we propose a new framework for zero-shot action recognition that explicitly disentangles motion and global information and enhances semantic alignment between visual and textual spaces. Firstly, we design a motion separation module to adaptively split video representations into global and dynamic components, thereby capturing fine-grained temporal cues that are often overlooked by prior work. Secondly, we introduce a motion aggregation block and a semantic guidance mechanism to highlight discriminative dynamic features while suppressing irrelevant background information. Thirdly, we incorporate negative prompts to enrich the textual space, providing stronger supervision for distinguishing unseen categories. Finally, we conduct extensive experiments on four public benchmarks, where our method consistently outperforms state-of-the-art approaches, demonstrating its effectiveness and strong generalization ability.

\vspace{-5pt}
\section{PROPOSED METHOD}
\label{sec:Method}
\begin{figure*}[ht]
  \centering
  \includegraphics[width=0.95\textwidth]{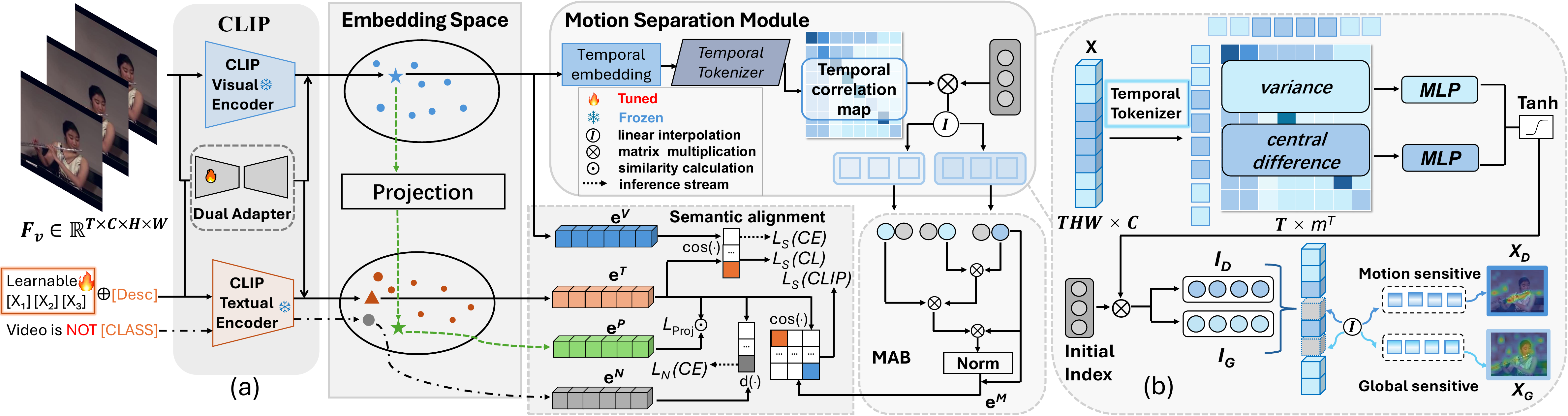}
  \vspace{-9pt}
  \caption{Overall pipeline of the proposed framework. (a) End-to-end architecture: video frames are processed by CLIP with a Dual Adapter (DA), followed by the Motion Separation Module (MSM), Motion Aggregation Block (MAB), and semantic alignment with textual prompts, where negative prompts expand the semantic space for better generalization. (b) MSM design: temporal features are tokenized, and variance and central-difference cues with learnable offsets are used to adaptively separate motion- and global-sensitive representations.}
  \vspace{-9pt}
  \label{fig:pipeline}
\end{figure*}

\vspace{-2pt}
\subsection{CLIP-based Video-Text Representation}
\vspace{-2pt}
\noindent
\noindent
As shown in Fig.~\ref{fig:pipeline}, given an input video 
$F_v \in \mathbb{R}^{T \times C \times H \times W}$ with $T$ frames, 
we first extract visual representations using a frozen CLIP visual encoder. 
To adapt CLIP to the video domain while preserving zero-shot generalization, 
we introduce a lightweight Dual Adapter (DA)~\cite{pfeiffer2020adapterhub}, 
which is shared across both visual and textual streams and applied in parallel 
to their encoders. 
Specifically, the adapter is inserted after every few frozen Transformer blocks 
and consists of a residual self-attention module followed by a bottleneck MLP, 
producing task-specific feature modulation while keeping the backbone frozen. 
By jointly adapting both visual and text encoders, the dual adapter effectively 
reduces cross-modal mismatch with minimal parameter overhead, enabling efficient 
video adaptation without sacrificing generalization.

For textual representations, instead of using only class names, we construct 
prompts that combine learnable context tokens with natural language descriptions 
of each action. Specifically, for the $k$-th class, the prompt is formulated as:
\begin{equation}
P_k = [p_k^1][p_k^2]\cdots[p_k^M]\,
[\text{Tokenizer(desc)}]_k
\end{equation}
where $[p_k^m]$ is the $m$-th learnable token, and $\text{Tokenizer}$ denotes the 
encoding of a class-specific description (e.g., ``a person playing the flute''). 
Feeding $P_k$ into the frozen CLIP text encoder $h(\cdot)$ yields the class 
embedding $f_k^T = h(P_k)$.

Inspired by \cite{ban2024understanding}, we adopt a \textit{Class-Dependent Negative Prompting} strategy to refine the semantic decision boundary.
Instead of using generic negations, we generate a class-specific negative description $T_{\text{neg}}^{k}$ (e.g., ``This video is NOT [CLASS$_k$]'') for each category $k$. These prompts are optimized using a multi-task contrastive objective that jointly increases the similarity between a video and its ground-truth positive prompt while decreasing the similarity to the corresponding class-specific negative prompt.
This push--pull mechanism explicitly enlarges the \textit{one-vs-rest} margin in the CLIP embedding space. By suppressing similarities to the most competitive classes, the model is encouraged to learn more discriminative representations for semantically similar actions (e.g., distinguishing \textit{BreastStroke} from \textit{FrontCrawl}), thereby providing stronger supervision for recognizing unseen categories.

\vspace{-5pt}
\subsection{Motion Separation Module (MSM)}
\vspace{-2pt}
Prior efforts either use fixed temporal splits~\cite{rodriguez2020deep} or enrich action semantics by intra-action decomposition, e.g., CLAVER~\cite{jingyi2025kronecker}. 
While effective for semantic diversity, these approaches do not explicitly localize motion nor disentangle it from static backgrounds, which often induce spurious correlations. 
Our Motion-guided Offset Learning (MOL) directly targets this issue by learning motion-aware temporal offsets, yielding a clean separation between motion-sensitive and global-static streams.

Given spatiotemporal features 
$X \in \mathbb{R}^{T  H  W \times C}$ with $T$ frames, we derive a per-frame embedding by spatial averaging:
\[
e^{(t)}=\tfrac{1}{HW}\sum_{h=1}^{H}\sum_{w=1}^{W} X_{t,h,w,:}, \quad t=1,\dots,T.
\]
Let $\mu=\tfrac{1}{T}\sum_{t=1}^{T} e^{(t)}$ denote the temporal mean descriptor. Two complementary motion statistics are then extracted. The deviation from the mean, 
\begin{equation}
v^{(t)}=\|e^{(t)}-\mu\|_2^2,
\end{equation}
measures whether a frame deviates from the overall distribution of the video, highlighting potential dynamic variations. The central difference,  
\begin{equation}
c^{(t)}=
\begin{cases}
\|e^{(2)}-e^{(1)}\|_2, & t=1,\\
\tfrac{1}{2}\|e^{(t+1)}-e^{(t-1)}\|_2, & 1<t<T,\\
\|e^{(T)}-e^{(T-1)}\|_2, & t=T,
\end{cases}
\end{equation}
captures local temporal change. Both statistics are normalized within each clip, and their weighted sum defines the motion saliency score: \begin{equation} m^{(t)}=\alpha \,\tilde v^{(t)}+\beta \,\tilde c^{(t)}, \end{equation} where $\alpha,\beta \ge 0$ are hyperparameters weighting the two terms. While $\tilde v^{(t)}$ denotes the normalized global deviation of frame $t$, and $\tilde c^{(t)}$ denotes the normalized local temporal change. To separate frames into two interleaved groups, we initialize global anchors $a_G(i)=2i-1$ and dynamic anchors $a_D(i)=2i$ for $i=1,\dots,\lfloor T/2\rfloor$. An offset generator $f_\theta$ (MLP followed by $\tanh$) maps saliency to bounded offsets: \begin{equation} \Delta_G^{(t)}=\delta\tanh\!\big(f_\theta^G(m^{(t)})\big), \quad \Delta_D^{(t)}=\delta\tanh\!\big(f_\theta^D(m^{(t)})\big), \end{equation} where $\delta>0$ is a hyperparameter controlling the maximum adjustment range, and $\Delta_G^{(t)}$, $\Delta_D^{(t)}$ denote the predicted temporal offsets for the global and dynamic anchors at frame $t$. The continuous sampling indices are 
\begin{equation}
\begin{aligned}
I_G(i) &= \min\!\big(\max(a_G(i)+\Delta_G^{(a_G(i))},\,1),\,T\big), \\
I_D(i) &= \min\!\big(\max(a_D(i)+\Delta_D^{(a_D(i))},\,1),\,T\big),
\end{aligned}
\end{equation}
where $I_G(i)$ and $I_D(i)$ denote the adjusted temporal 
sampling positions for the global and dynamic groups, respectively.

Because these indices may be fractional, we apply differentiable linear interpolation 
on the original video tensor $X \in \mathbb{R}^{T\times H\times W\times C}$:
\begin{align}
X_G[i] &= (1-\gamma(I_G[i]))X_{\lfloor I_G[i]\rfloor,:,:,:}
        + \gamma(I_G[i])X_{\lceil I_G[i]\rceil,:,:,:}, \\
X_D[i] &= (1-\gamma(I_D[i]))X_{\lfloor I_D[i]\rfloor,:,:,:}
        + \gamma(I_D[i])X_{\lceil I_D[i]\rceil,:,:,:},
\end{align}
where $X_G[i]$ and $X_D[i]$ denote the interpolated frame features sampled at 
positions $I_G(i)$ and $I_D(i)$, respectively, and 
$\gamma(\tau)=\tau-\lfloor\tau\rfloor$ is the fractional part 
(with $\lfloor\cdot\rfloor$ and $\lceil\cdot\rceil$ denoting floor and ceil operators).

In this way, MSM leverages motion-aware offsets to adaptively separate global/static features $X_G$ and motion-sensitive features $X_D$. The design ensures that frames with high deviation or strong local change are routed to the dynamic stream, while more stable frames contribute to the global stream. This learned separation provides finer supervision for action modeling and yields more robust representations for zero-shot recognition.

\vspace{-2pt}
\subsection{Motion Aggregation Block (MAB)}
\vspace{-2pt}
After obtaining the motion-sensitive features $X_D$ and global-sensitive features $X_G$ from the motion-guided offset learning module, we employ a Motion Aggregation Block (MAB) to integrate these complementary sources.
Specifically, $X_D$ encodes fine-grained temporal variations, while $X_G$ preserves stable contextual patterns.

To bridge these streams, independent reweighting alone is insufficient for modeling entangled actions, as it ignores the dependency between motion and context.
Therefore, MAB employs a gated interaction mechanism designed to explicitly model \textit{motion--appearance co-occurrence}.
Formally, the fused representation $e^{M}$ is computed as:
\begin{equation}
    e^{M} = \text{LayerNorm}\left( X_D + \sigma\left( \mathbf{W}_g \left[ (X_D \odot X_G), X_D, X_G \right] \right) \odot X_G \right),
\end{equation}

\noindent where $\odot$ denotes element-wise multiplication, $[\cdot]$ represents feature concatenation along the channel dimension, $\sigma(\cdot)$ is the sigmoid function, and $\mathbf{W}_g$ denotes learnable gating weights.
The explicit interaction term $(X_D \odot X_G)$ enables motion-conditioned filtering of global context, allowing MAB to selectively enhance motion-relevant semantics while preserving original motion cues through the residual connection. This design enables MAB to selectively emphasize motion dynamics while 
retaining complementary global semantics, producing stable embeddings for downstream classification.

\vspace{-2pt}
\subsection{Multi-Objective Semantic Alignment}
\vspace{-2pt}
To optimize our framework, we design a multi-objective learning scheme consisting of two major parts: seen class alignment loss $\mathcal{L}_{S}$ and negative prompt loss $\mathcal{L}_{N}$. The overall objective is written as
\begin{equation}
\mathcal{L} = \mathcal{L}_{S} + \lambda_{N}\mathcal{L}_{N},
\end{equation}
where $\lambda_{N}=0.1$ is empirically set based on our experiments.

\noindent\textbf{Seen class alignment.}  
Given video embeddings $e^V$ and textual embeddings $e^T$, we follow CLIP to obtain predictions:
\begin{equation}
P_{S,i} = \mathrm{softmax}(\cos(e^V_i, e^T)),
\end{equation}
where $\cos(\cdot)$ denotes cosine similarity. The seen class alignment loss $\mathcal{L}_{S}$ integrates several complementary objectives:
\begin{equation}
\label{eq10}
\begin{aligned}
\mathcal{L}_{S} = & \sum_i \mathrm{CE}(P_{S,i}, y_i) 
+ \mathcal{L}_{CL}^{S}(e^V_i, e^T) \\
& + \mathcal{L}_{CLIP}^{S}(e^V_i, e^T_{\text{CLIP}}) 
+ \mathcal{L}_{Proj} .
\end{aligned}
\end{equation}

Here, the first term is a cross-entropy loss ensuring that each video aligns with its ground-truth class.  
$\mathcal{L}_{CL}^{S}$ is a contrastive loss that enlarges inter-class margins by pulling positive pairs closer and pushing negatives apart.  
$\mathcal{L}_{CLIP}^{S}$ preserves consistency with the pretrained CLIP embedding space, avoiding domain drift during adaptation.  
Finally, $\mathcal{L}_{Proj}$ is a projection loss defined as
\begin{equation}
\mathcal{L}_{Proj} = \| e^P - e^T \|_2^2,
\end{equation}
which regularizes positive prompt embeddings $e^P$ to remain close to the original class semantics $e^T$.

\noindent\textbf{Unseen class regularization.}  
For unseen categories, we adopt class-dependent negative prompts (``video is not [CLASS]’’). Given negative embeddings $e^N$, we compute
\begin{equation}
P_{N,i} = \mathrm{softmax}(\cos(e^V_i, e^N)),
\end{equation}
and define
\begin{equation}
\mathcal{L}_{N} = \sum_i \mathrm{CE}(P_{N,i}, \hat{y}_i),
\end{equation}
where $\hat{y}_i$ denotes the complementary “not-class” label. This loss forces the model to de-emphasize spurious correlations, thereby improving generalization in zero-shot scenarios.

\begin{table}[ht]
\centering
\scriptsize
\caption{Zero-shot performance on HMDB-51, UCF-101, and K-600. 
All results are reported in accuracy (\%).}
\label{tab:hmdb-ucf-k600}

\scalebox{0.9}{
\begin{tabular}{l l ccc}
\toprule
\textbf{Method} & \textbf{Publication} & \textbf{HMDB-51} & \textbf{UCF-101} & \textbf{K-600} \\
\midrule
\multicolumn{5}{l}{\textit{Methods with Vision Training}} \\
ER-ZSAR~\cite{chen2021elaborative} & ICCV'21 & 35.3 $\pm$ 4.6 & 51.8 $\pm$ 2.9  & 42.1$\pm$1.4 \\
JigSawNet~\cite{le2019jigsawnet}   & TIP’19 & 39.3 $\pm$ 3.9 & 56.8 $\pm$ 2.8  & --- \\
\midrule
\addlinespace[2pt]
\multicolumn{5}{l}{\textit{Methods with Vision-Language Training}} \\
A5~\cite{ju2022prompting}  & ECCV'22 & 44.3 $\pm$ 2.2 & 69.3 $\pm$ 4.2  & --- \\
X-CLIP~\cite{ni2022expanding}          & ECCV'22   & 46.3 $\pm$ 0.6 & 70.3 $\pm$ 2.3  & 67.1$\pm$1.0 \\
Vita-CLIP~\cite{wasim2023vita}         & CVPR'23   & 48.6 $\pm$ 0.6 & 75.0 $\pm$ 0.6  & --- \\
SDR-CLIP~\cite{gowda2024telling}       & ACCV'24   & 52.7 $\pm$ 1.4 & 75.3 $\pm$ 3.2  & --- \\
GIL~\cite{gowda2024continual} & ACCV'24   & 53.9 $\pm$ 1.4 & 79.4 $\pm$ 1.4  & --- \\
TP-CLIP~\cite{gowda2025temporal}       & CVPR'25   & 54.1 $\pm$ 1.2 & 81.1 $\pm$ 1.2  & --- \\
CLAVER~\cite{jingyi2025kronecker} & ICLR'25 & 54.1$\pm$2.4 & 78.6 $\pm$ 1.7  & --- \\
STDD~\cite{yu2025building}             & AAAI'25   & 55.9$\pm$0.2 & 85.2 $\pm$ 1.2  & 75.1$\pm$ 0.7 \\
ZAR~\cite{liang2025zar}                & EI'25   & 54.2$\pm$0.8 & 77.4 $\pm$ 0.8  & 70.5$\pm$ 0.4 \\
\rowcolor{oursblue}\textbf{Ours}                          & ---       & \textbf{55.2$\pm$1.1} & \textbf{82.2 $\pm$ 0.5} & \textbf{72.4$\pm$ 0.5} \\
\bottomrule
\end{tabular}}
\end{table}

\vspace{-6pt}

\begin{table}[t]
\centering
\scriptsize

\caption{Base-to-novel generalization on Kinetics400 and HMDB-51. HM = harmonic mean of Base and Novel. Accuracy in \%.}
\scalebox{0.9}{\begin{tabular}{l|ccc|ccc}
\toprule
& \multicolumn{3}{c|}{\textbf{Kinetics400}} & \multicolumn{3}{c}{\textbf{HMDB-51}} \\
\midrule
\textbf{Method} & Base & Novel & HM & Base & Novel & HM \\
\midrule
Vanilla CLIP B/16~\cite{radford2021learning} & 53.3 & 46.8 & 49.8 & 53.3 & 46.8 & 49.8 \\
ActionCLIP B/16~\cite{wang2023actionclip}    & 69.0 & 57.2 & 62.6 & 69.1 & 37.3 & 48.5 \\
XCLIP B/16~\cite{ni2022expanding}            & 74.1 & 56.4 & 64.0 & 69.4 & 45.5 & 55.0 \\
A5~\cite{ju2022prompting}        & 74.1 & 56.4 & 64.0 & 46.2 & 16.0 & 23.8 \\
ViFi-CLIP B/16~\cite{rasheed2023fine}        & 76.4 & \textbf{61.1} & 67.9 & 73.8 & 53.3 & 61.9 \\
ZAR B/16~\cite{liang2025zar}                 & 75.2 & 60.7 & 67.2 & 75.2 & 55.2 & 63.7 \\
\rowcolor{oursblue}\textbf{Ours}                                 & \textbf{78.8} & 60.6 & \textbf{68.5} & \textbf{78.5} & \textbf{60.4} & \textbf{68.3} \\
\bottomrule
\end{tabular}}
\label{tab:base2novel}
\end{table}

\begin{table}[htbp]
\centering
\scriptsize
\caption{Ablation study across three modules and three datasets. All results are reported in accuracy (\%).}
\label{tab:ablation_modules}
\scalebox{0.9}{
\begin{tabular}{cccc|ccc}
\toprule
\multicolumn{4}{c|}{\textbf{Component / Setting}} & \multicolumn{3}{c}{\textbf{Zero-shot (\%)}} \\
\textbf{DA} & \textbf{MSM} & \textbf{MAB} & \textbf{Splitting} & \textbf{HMDB-51} & \textbf{UCF-101} & \textbf{SSv2} \\
\midrule
\multicolumn{4}{c|}{Baseline (Only CLIP)}                  & 36.3 & 58.3 & 5.1 \\
\checkmark &             &             & Offsets &  (+1.1\%)  &  (+0.5\%) &  (+41.2\%) \\
           & \checkmark  &             & Offsets & (-0.8\%)  &  (+3.4\%) &  (+27.5\%) \\
 &      \checkmark       & \checkmark  & Offsets & (+11.0\%) &  (+6.3\%) &  (+56.9\%) \\
           
\rowcolor{oursblue}\checkmark & \checkmark  & \checkmark  & Offsets & \textbf{ (+23.1\%)} & \textbf{ (+13.4\%)} & \textbf{(+62.7\%)} \\
\midrule

           & \checkmark  &    \checkmark          & Fixed   & (+13.3\%) & (+1.2\%) & (+15.3\%) \\
\bottomrule
\end{tabular}}
\end{table}

\begin{table}[t]
\centering
\scriptsize
\caption{Effects of different losses. Checkmarks indicate the losses used. All results are reported in accuracy (\%).}
\label{tab:loss_effect}
\scalebox{0.9}{
\begin{tabular}{ccccccc}
\toprule
$\mathcal{L}_{CE}^{S}$ & $\mathcal{L}_{CL}^{S}$ & $\mathcal{L}_{CE}^{N}$ & $\mathcal{L}_{CLIP}^{S}$ & $\mathcal{L}_{Proj}$ & HMDB & UCF \\
\midrule
\checkmark & \blank & \blank & \blank & \blank & 46.0 & 71.3 \\
\checkmark & \blank & \checkmark & \blank & \blank & 54.8 & 80.5 \\
\blank & \blank & \blank & \checkmark & \blank & 53.5 & 78.8 \\
\checkmark & \checkmark & \checkmark & \checkmark & \blank & 55.0 & 81.7 \\
\rowcolor{oursblue}\checkmark & \checkmark & \checkmark & \checkmark & \checkmark & \textbf{55.2} & \textbf{82.2} \\
\bottomrule
\end{tabular}}
\end{table}

\begin{figure}[htbp]
    \centering
   
   \includegraphics[width=\columnwidth]{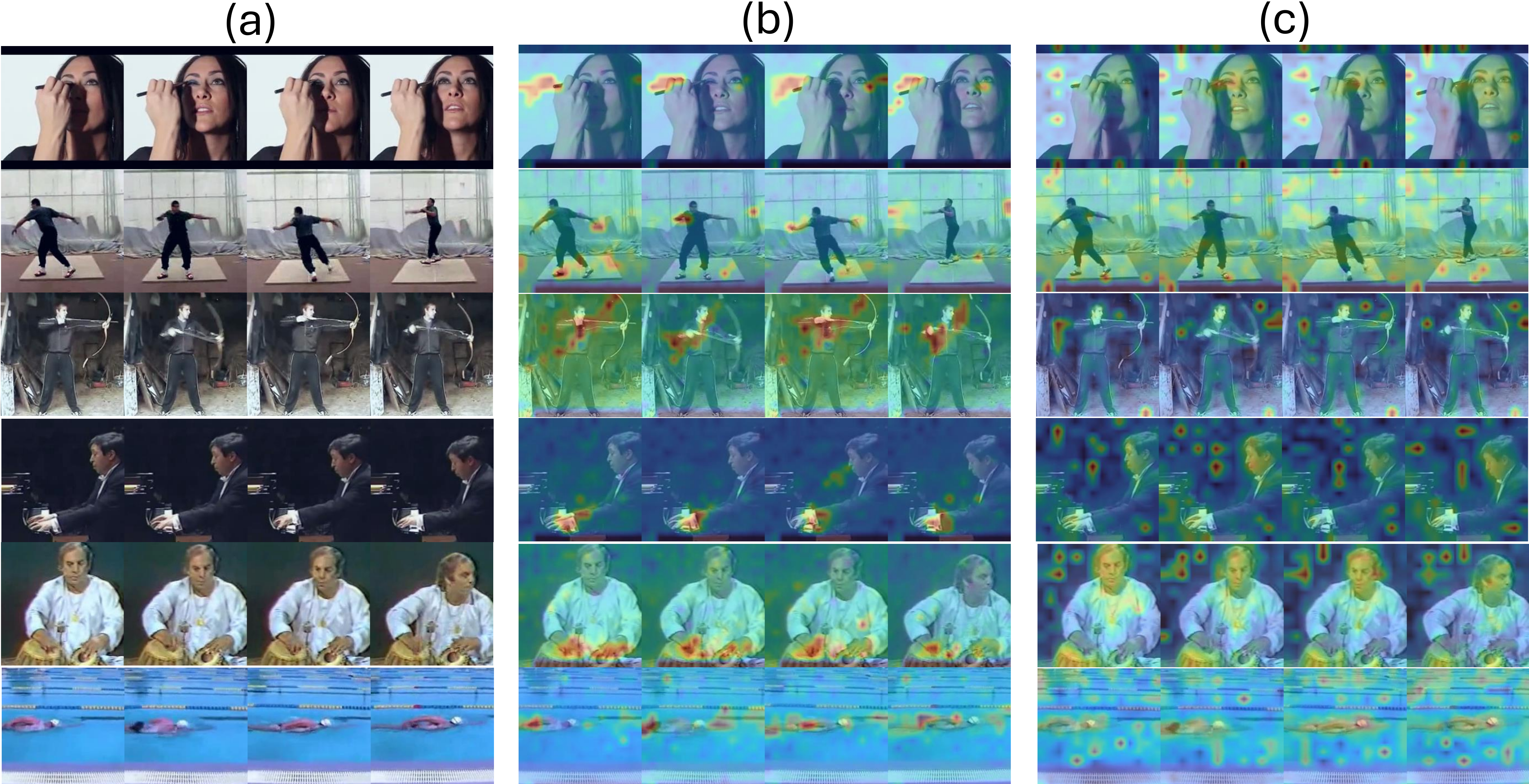}
   \vspace{-10pt}
    \caption{Attention visualization on HMDB-51. (a) shows raw video frames, (b) highlights motion-sensitive regions, and (c) emphasizes global-static patterns.}
    \vspace{-10pt}
    \label{fig:att}
\end{figure}

\vspace{-6pt}
\section{Experiments}
\label{sec:pagestyle}
\vspace{-2pt}
\subsection{Experimental Setup}
\vspace{-2pt}
To evaluate our method, we conduct experiments on five widely used benchmarks: Kinetics-400~\cite{zisserman2017kinetics}, Kinetics-600~\cite{carreira2018short}, HMDB51~\cite{kuehne2011hmdb}, UCF101~\cite{soomro2012ucf101}, and Something-Something V2 (SSv2)~\cite{goyal2017something}. Kinetics-400 serves as the training set, while HMDB51, UCF101, SSv2, and Kinetics-600 (excluding overlaps with K400) are used for zero-shot evaluation. HMDB51 and UCF101 are standard action recognition benchmarks, whereas SSv2 emphasizes fine-grained temporal reasoning. Our backbone is ViT-B/16 CLIP, trained with sparse frame sampling (16 or 32 frames). We adopt AdamW with a learning rate of $1 \times 10^{-5}$, cosine decay, and batch size 64 for 100 epochs on an NVIDIA A100 80GB.

\vspace{-2pt}
\subsection{Zero-shot Performance}
\vspace{-2pt}
In Table~\ref{tab:hmdb-ucf-k600}, The zero-shot performance on HMDB-51, UCF-101, and Kinetics-600 in reported. Our method achieves 55.2\% on HMDB-51 and 82.2\% on UCF-101, surpassing recent vision–language approaches such as STDD and TP-CLIP, while obtaining 72.4\% on K-600, which is competitive with the best existing result. These improvements primarily arise from disentangling motion-sensitive and global-sensitive cues via the MSM and fusing them through gated interactions in the MAB, which reduces background bias and preserves fine-grained temporal patterns. The gains are most evident on motion-centric actions where appearance is ambiguous (e.g., clapping, waving), while failure cases often occur in subtle object-centric activities or highly scene-dependent classes (e.g., swimming vs.\ surfing). Overall, the results demonstrate that our approach significantly enhances the generalization ability of CLIP to unseen actions in the zero-shot setting.
\vspace{-2pt}
\subsection{Base-to-novel performance}
\vspace{-2pt}
The base-to-novel setting provides a principled protocol to assess a model’s generalization to unseen categories, analogous to the zero-shot learning scenario. As shown in Table~\ref{tab:base2novel}, the vanilla CLIP~\cite{radford2021learning} baseline yields the weakest performance, confirming the large gap between image-pretrained models and video action recognition. ActionCLIP~\cite{wang2023actionclip} and X-CLIP~\cite{ni2022expanding} improve the base accuracy but suffer from limited generalization to novel classes, resulting in moderate HM scores. ViFi-CLIP~\cite{rasheed2023fine} achieves the highest novel accuracy (61.1\%), yet its base accuracy is lower than ours. In contrast, our method attains the best base accuracy (78.8\%) while maintaining competitive novel performance (60.6\%), leading to the highest harmonic mean (68.5\%). This demonstrates that our approach achieves a more favorable trade-off between base and novel classes, ensuring both strong discriminability on seen categories and improved generalization to unseen ones.

\vspace{-2pt}
\subsection{Ablation Study}
\vspace{-2pt}
We evaluate each component under the vanilla CLIP setting (Tab.~\ref{tab:ablation_modules}). 
The Dual Adapter (DA), as a lightweight parallel module, adapts both image and text features and delivers clear gains with few trainable parameters. MSM alone slightly reduces HMDB performance (-0.8\%) due to frozen CLIP features, but when combined with DA, it brings large improvements (+11.0\% on HMDB and +56.9\% on SSv2). This shows DA is essential for bridging the image–video gap, enabling temporal alignment, and preserving zero-shot transferability. Adding MAB on top of DA+MSM nearly doubles the improvements. On HMDB-51, the fixed splitting manner slightly outperforms offsets. We attribute this to coarse-grained nature, where temporal precision is less critical and the fixed rule offers a more stable partition. In contrast, on larger and more temporally demanding datasets, the adaptive offsets consistently deliver superior gains.

For semantic alignment, we design multiple losses. 
As shown in Table~\ref{tab:loss_effect}, adding the negative prompt loss $\mathcal{L}_{CE}^{N}$ yields over 10\% gains on both datasets, demonstrating the benefit of leveraging unseen negative representations. 
The comparison between the fourth and last rows further confirms the necessity of combining $\mathcal{L}_{S}$ and $\mathcal{L}_{N}$. Our framework thus achieves classification alignment by enlarging inter-class margins while preserving CLIP’s semantic consistency: positive prompts align with original semantics, while negative prompts suppress background noise and provide non-class textual cues to enhance adaptation to unseen categories. 
Overall, the full objective secures discriminability on seen classes and strong zero-shot generalization across datasets.
\vspace{-2pt}
\subsection{Visualization}
\begin{figure}[htbp]
    \centering
    
    \includegraphics[width=\columnwidth]{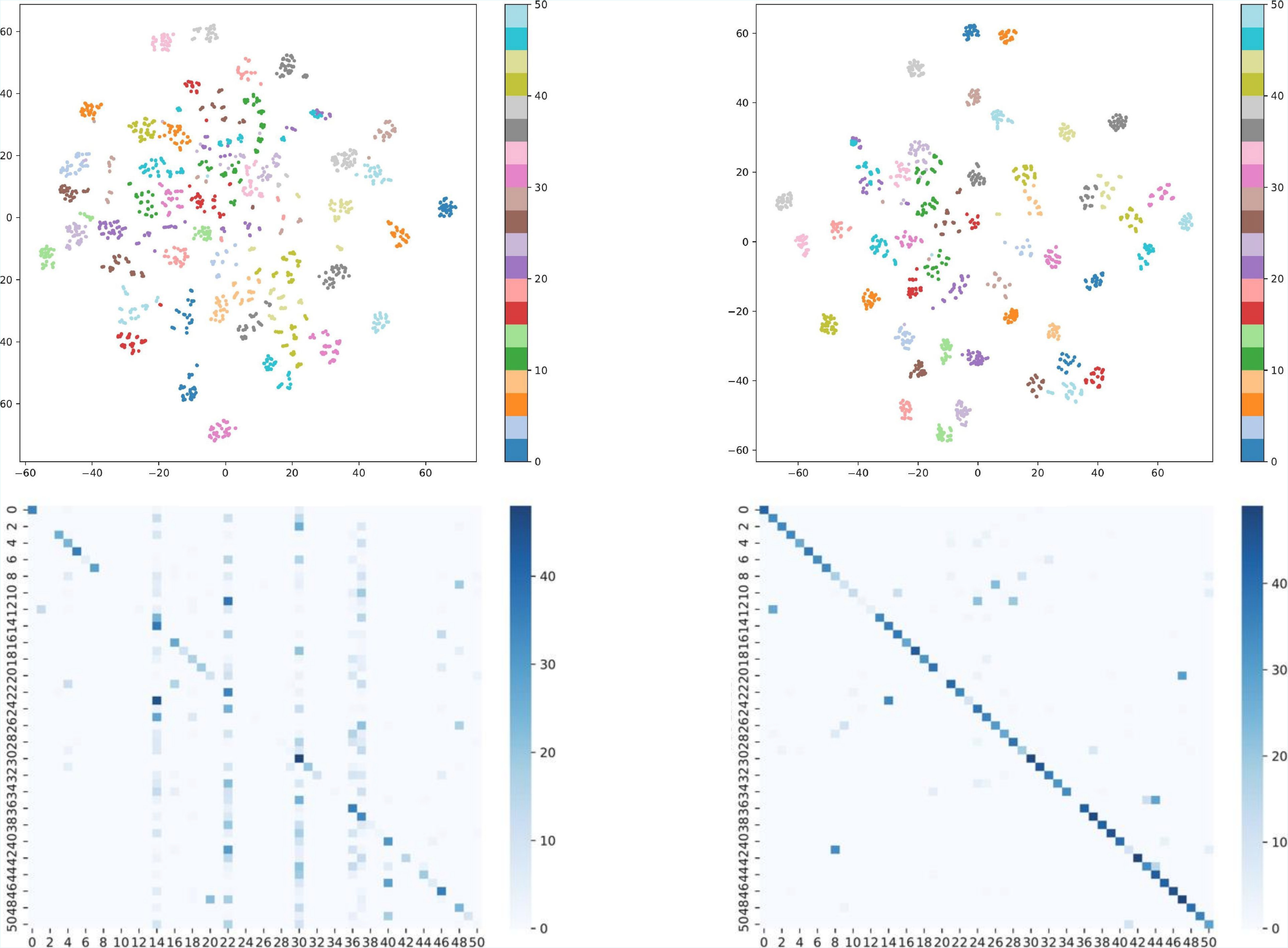}
    \vspace{-10pt}
    \caption{t-SNE visualizations (top) and confusion matrices (bottom) on HMDB-51. The color bars on the right denote the 51 actions.}
\vspace{-10pt}
    \label{fig:vis}
\end{figure}

\vspace{-2pt}
Fig.~\ref{fig:att} shows attention map visualizations of temporal dynamics and global static cues. The static branch consistently highlights body regions, while the dynamic branch focuses on action-relevant areas, suppressing distractions and background noise. Both are crucial for accurate video action understanding. Moreover, in the second row, the motion attention maps successfully capture the moving object—the discus itself. In Fig.~\ref{fig:vis}, compared with Vanilla CLIP (left), our method (right) produces more compact and separable clusters and a cleaner confusion matrix with stronger diagonal dominance, indicating improved discriminability and reduced class confusion.


\vspace{-6pt}
\section{Conclusion}
\vspace{-2pt}
\label{sec:conclusion}
    In conclusion, our motion-guided framework effectively disentangles motion and global cues, integrates them into semantically aligned representations, and leverages negative prompts for robust learning. These designs jointly enhance the model’s ability to generalize from base to novel classes, providing a principled step toward zero-shot video action recognition. Comprehensive experiments confirm the superior effectiveness of our framework and its strong adaptability to unseen categories.

\vfill\pagebreak

\bibliographystyle{IEEEbib}
\bibliography{strings,refs}

\end{document}